\pdfoutput=1
\documentclass[11pt]{article}
\usepackage[font=libertinus,citestyle=numeric]{kurbanlab}
\DeclareAffiliation{tamu}{Department of Electrical and Computer Engineering, Texas A\&M University, College Station, Texas, USA}
\DeclareAffiliation{ankara}{Department of Prosthetics and Orthotics, Ankara University, Ankara, Turkey}
\DeclareAffiliation{tamuq}{Department of Electrical and Computer Engineering, Texas A\&M University at Qatar, Doha, Qatar}
\DeclareAffiliation{hbku}{College of Science and Engineering, Hamad Bin Khalifa University, Doha, Qatar}

\graphicspath{{figures/}}

\newcommand{\angstrom}{\text{\AA}}

\title{Separating perception from reasoning in vision-language models: a model-free render ceiling for crystal structures}
\RunningTitle{A model-free render ceiling for crystal structures}
\Author{Can Polat}{tamu}
\Author[corresponding=kurbanm@ankara.edu.tr]{Mustafa Kurban}{ankara,tamuq}
\Author{Erchin Serpedin}{tamu}
\Author[corresponding=hkurban@hbku.edu.qa]{Hasan Kurban}{hbku}
\begin{document}
\maketitle
\begin{abstract}
Vision-language models (VLMs) often predict plausible structural labels from images without recovering the three-dimensional geometry behind them. End-to-end scores cannot tell whether an error comes from ambiguous views, lost visual information, or flawed reasoning. Multi-view 3D reconstruction faces its own trade-off: rigid symmetry priors erase genuine distortions, while unconstrained triangulation absorbs measurement outliers. We separate visual perception from structural reasoning through an inspectable 3D geometric intermediate. Its foundation is a model-free \emph{render ceiling}: an exact reference for what calibrated, species-labeled projections determine, obtained by ruling out projective phantoms. Across a 1,950-structure census, every tested combination of four or five calibrated views uniquely determined the 3D sites from complete, species-labeled centroids. Against this reference, replacing images with exact text coordinates lowers the error rate of all 17 evaluated VLMs on crystals. The three frontier models gain 10.0--19.5 percentage points yet still miss 8.1--21.0\% of cases. To recover geometry from imperfect observations, our symmetry-aware split reconstruction jointly estimates a symmetric base, coherent 3D physical displacements, and view-specific 2D corruption, adaptively selecting between atomic and collective distortion modes. Relative to Huber triangulation, it reduces coordinate error by 36.7\% under Gaussian noise and 41.2\% under corruption. Adaptive mode selection further cuts collective error up to 30.4\% relative to atomic split across two external cohorts without sacrificing localized fidelity. Given reconstructed rather than Huber coordinates, downstream models improve. Geometric recovery thus becomes an explicit, testable foundation for structural understanding.
\end{abstract}
\section{Introduction}
\label{sec:intro}
Vision-language models (VLMs) struggle to extract precise geometric relationships from images~\cite{kamoi2025visonlyqa} and routinely overlook spatial structure already encoded within their visual representations~\cite{fu2025hidden}. While synthetic benchmarks~\cite{johnson2017clevr}, controllable scene generators~\cite{greff2022kubric}, and neuro-symbolic architectures~\cite{amizadeh2020neurosymbolic} provide structured tasks and intermediates for everyday visual reasoning, scientific interpretation requires knowing whether a prediction is supported by faithful three-dimensional geometry. Scoring end-to-end labels alone cannot distinguish information absent from the 2D observations, geometry lost during perception, and structural evidence supplied explicitly but used incorrectly~\cite{zhang2024mathverse}.

Crystal structures make this diagnostic blind spot concrete. Atomic electron tomography resolves 3D atomic arrangements and subtle surface displacements~\cite{lee2021atomic}, yet projection misalignment degrades reconstruction fidelity~\cite{wang2022alignment}. Even noiseless directional projections can admit projective phantoms---false 3D point sets whose 2D projections coincide with real atoms in every view~\cite{gardner1997tomography}. An incorrect symmetry label can therefore reflect insufficient views or flawed visual processing, whereas a correct label leaves individual atomic coordinates unverified~\cite{spek2009structure}. Evaluating structural understanding requires both a reference for what the observations determine and a direct assessment of the recovered 3D geometry~\cite{el2024probing}.

Under imperfect observations, recovering that geometry exposes a fundamental trade-off between preserving physical distortions and rejecting measurement error. Crystallographic analysis under noise must separate genuine symmetry breaking from pseudosymmetry~\cite{moeck2022symmetry} while capturing localized shifts or symmetry-adapted collective distortion modes~\cite{campbell2006isodisplace}. Existing approaches are caught between two extremes: enforcing a rigid symmetry prior suppresses true physical departures, whereas unconstrained triangulation absorbs single-view outliers into false 3D displacements~\cite{wu2020unsupervised}. The unresolved question is how to preserve both localized and collective departures while rejecting corruption when neither the parent symmetry nor the displacement representation is supplied as ground truth~\cite{sim2006removing}.

Here, using five calibrated orthographic views, complete species-labelled centroids, and a known lattice, we establish multi-view recoverability, reconstruct noisy coordinates, and test a 112-case handoff of reconstructed coordinates to three downstream models. Our contributions are:
\begin{itemize}[leftmargin=*,itemsep=2pt]
\item \textbf{A model-free render ceiling.} We enumerate the 3D sites compatible with calibrated, species-labeled projections and verify across the census that every true site is required. The ceiling serves as a reference for what the views themselves support, with a finite-tolerance implementation for empirical use.
\item \textbf{A paired diagnostic of sensitivity to supplied geometry.} Holding the model, structure, and question fixed, we measure which image-input errors disappear when exact geometry is supplied, which persist, and which new errors appear.
\item \textbf{Symmetry-aware split reconstruction.} From noisy or corrupted projections, we jointly estimate an inferred crystallographic base, coherent 3D physical departures, and view-specific observation errors. On the reserved cohort, our method reduces coordinate RMSD by 36.7\% under Gaussian noise and 41.2\% under single-view corruption relative to Huber triangulation. Across two external cohorts, adaptive selection between atomic and collective modes further reduces collective displacement error by 27.4--30.4\% relative to atomic split.
\end{itemize}
\section{Related work}
\label{sec:related}
\textbf{Disentangling visual perception from reasoning.} Controlled visual probing reveals persistent perceptual blind spots in multimodal models~\cite{tong2024eyeswideshut}, motivating structured intermediates that make each stage of visual reasoning inspectable from symbolic scene parsing in NS-VQA~\cite{yi2018nsvqa} and vector-graphics primitives in VDLM~\cite{vdlm2404} to perceptual descriptions that expose perception bottlenecks in abstract-reasoning benchmarks~\cite{arc2512}. Recent post-training frameworks similarly decouple visual perception, visual reasoning, and textual reasoning through staged optimization~\cite{decouple2605} or visual-dependence reward normalization in PDCR~\cite{pdcr2026}. While crystallographic benchmarks have extended multimodal evaluation to scientific structures and composition exclusions~\cite{xcrys2506}, existing evaluations still score end-to-end symmetry labels rather than testing whether the underlying 3D coordinates are uniquely determined by the views or faithfully recovered prior to reasoning~\cite{polat2026scalar}.

\textbf{Multi-view 3D reconstruction under measurement outliers.} Classical multi-view geometry recovers 3D shape from tracked orthographic projections via factorization~\cite{tomasi1992factorization} or from matched image observations via triangulation~\cite{hartley1997triangulation}, while All-Angles Bench evaluates cross-view correspondence and geometric understanding in modern multimodal models~\cite{allangles2504}. To handle corrupted observations, robust multiview inverse methods incorporate sparse outlier estimation~\cite{dalalyan2012robust} or uncertainty-aware triangulation~\cite{lee2020robust}.

\textbf{Balancing symmetry priors against structural departures.} Incorporating symmetry without over-constraining the recovered structure is a central tension in both 3D vision and atomic tomography. SNeS applies soft symmetry priors to neural surface geometry and materials while factoring out asymmetric appearance~\cite{insafutdinov2022snes}, whereas NSIRE reconstructs atomic-potential volumes by combining a diffusion prior with null-space updates that enforce tomographic projection consistency~\cite{li2026nsire}. For discrete crystal sites, the corresponding question is whether unordered 2D centroids support a candidate space group and which localized or collective departures should be retained.

\textbf{Structured sparse decompositions and model selection.} Separating superimposed structural components is well established in high-dimensional estimation: dirty models decompose parameters across distinct regularizers~\cite{jalali2010dirty}, and robust group lasso jointly estimates a group-sparse signal alongside sparse measurement corruption~\cite{wei2015robust}. Cross-validation can itself overfit when selecting among competing structural hypotheses~\cite{cawley2010selection}, making the selection procedure and its evaluation relevant alongside the decomposition.

\section{Geometric recoverability, reconstruction, and diagnosis}
\label{sec:method}

\begin{figure*}[t]
\centering
\includegraphics[width=\textwidth]{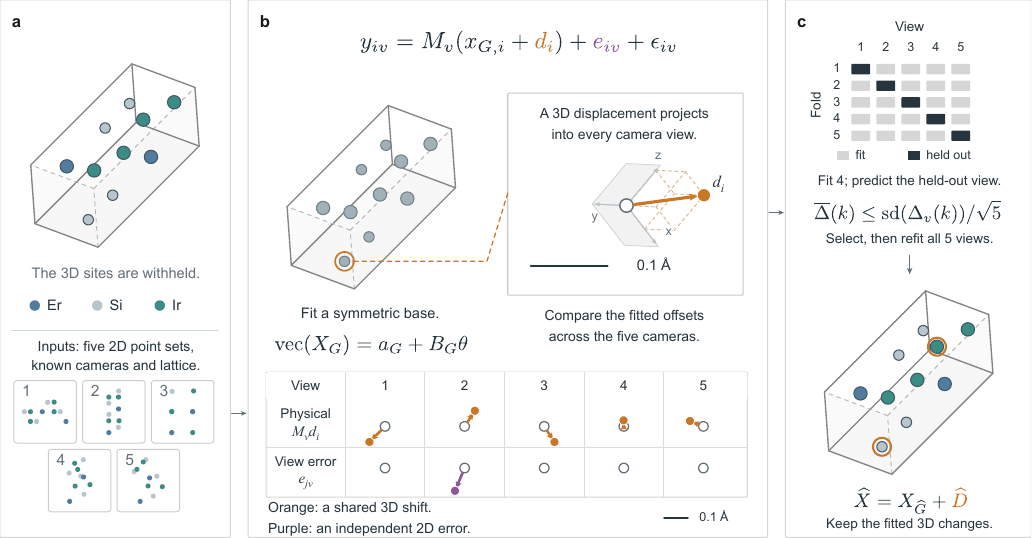}
\caption{\textbf{Separating coherent 3D physical departures from view-specific observation errors preserves symmetry-breaking structure.} \textbf{(a)} Five unordered, species-labeled projections provide inputs with known cameras and lattice; true sites and cross-view identities are withheld. \textbf{(b)} The split model separates a symmetry-consistent base ($X_G$, gray circles), shared 3D physical departures ($D$, orange rings), and view-specific observation errors ($E$). \textbf{(c)} Withheld-view prediction selects the penalty by paired one-SE, followed by an unpenalized all-view support refit and symmetry classification, shown on the held-out structure Er$_2$Si$_4$Ir$_4$ under Gaussian noise and one corrupted centroid (full cells share viewpoint and scale; inset and 2D grid have separate scale bars).}
\label{fig:method}
\end{figure*}

To decouple visual perception from crystallographic reasoning, we evaluate the 3D geometric intermediate across three regimes. Complete calibrated projections define an exact recoverability reference. Noisy or corrupted centroids test symmetry-aware split reconstruction across displacement representations and selection rules. Paired VLM interventions test both reference geometry and reconstructed coordinates as downstream inputs.

\subsection{When do the views determine a structure?}
A crystal is defined by a lattice $L$, Cartesian cell sites $P=\{p_i\}_{i=1}^{N}$, and species $Z=\{z_i\}_{i=1}^{N}$, following pymatgen~\cite{ong2013pymatgen}. Given known orthographic cameras $M_v\in\mathbb R^{2\times3}$ with orthonormal rows, ideal observations are unordered species-labeled sets $Q_{v,z}=\{M_vp_i:z_i=z\}$ that retain all centroids (including sites occluded in raster images) while withholding cross-view atom identities. Because any candidate site of species $z$ must project onto an observed centroid of the same species in every view $v\in\mathcal C$, intersecting the back-projected rays defines the compatible set $A_z(\mathcal C)$, and subtracting the true sites $P_z$ isolates projective phantoms $\Phi_z$:
\begin{equation}
\label{eq:compatibility}
A_z(\mathcal C)=\bigcap_{v\in\mathcal C}M_v^{-1}(Q_{v,z}),\qquad \Phi_z=A_z(\mathcal C)\setminus P_z.
\end{equation}
Fix two views in $\mathcal C$ with nonparallel projection directions and call them anchors.

\begin{theorem}[Exact geometric recovery]
\label{thm:exact}
Given exact species-labeled projections, intersecting every pair of same-species anchor rays and keeping each intersection that projects onto an observed centroid in all remaining views yields exactly $A_z(\mathcal C)$. If $\Phi_z$ is empty for every species, removing duplicates recovers the true point sets, and hence any deterministic label computed with the supplied lattice.
\end{theorem}
\begin{proof}
Each true point lies on one ray from each anchor and satisfies every remaining projection constraint, so it is accepted. Conversely, every accepted point satisfies the defining constraints of $A_z$. Because the anchor directions are nonparallel, each pair of anchor rays meets in at most one point, so enumerating all same-species pairs produces every point of $A_z$ and the procedure is exhaustive. If $\Phi_z=\emptyset$, then $A_z=P_z$.
\end{proof}

Adding views monotonically shrinks or preserves $A_z(\mathcal C)$. Our finite-tolerance implementation permits closest-approach ray midpoints, enforces ray and remaining-view projection distances at most $\delta$, and suppresses later same-species candidates within $\delta$ of a retained point. We score its output against spglib~\cite{spglib2024} reference labels (positional tolerance $0.01\,\angstrom$, angle tolerance $5^\circ$) and denote the resulting label accuracy $R_{\mathrm{geo}}(\delta)$, which serves as the geometric reference for the paired diagnostic. We report $R_{\mathrm{geo}}$ at $\delta=0.01\,\angstrom$ and at the $0.15\,\angstrom$ tolerance used in the released benchmark.

\subsection{Separating structural and observation departures}
Under imperfect observations, reconstruction must distinguish a physical 3D shift---which produces linked offsets across all cameras---from an observation error isolated to a single view (Figure~\ref{fig:method}a--c). Given tentative correspondences for a candidate space group $G$, we decompose atom $i$'s observation in view $v$ into a symmetry-consistent base site $x_{G,i}\in\mathbb R^3$, a physical departure $d_i\in\mathbb R^3$, a view-specific error $e_{iv}\in\mathbb R^2$, and background noise $\epsilon_{iv}$:
\begin{equation}
y_{iv}=M_v(x_{G,i}+d_i)+e_{iv}+\epsilon_{iv}.
\label{eq:split}
\end{equation}
Because the reconstructed structure in~\eqref{eq:split} is $X_G+D$, its recomputed label can freely depart from the parent symmetry $G$.

Writing the candidate symmetric base as an affine space $\operatorname{vec}(X_G)=a_G+B_G\theta$ (from periodic symmetry operations and species-preserving permutations with a free origin), we balance data fidelity against sparse departures weighted by the camera information matrix $H=\sum_v M_v^{\mathsf T}M_v$:
\begin{align}
\min_{\theta,D,E}\;&\frac12\sum_{i,v}\|y_{iv}-M_v(x_{G,i}+d_i)-e_{iv}\|_2^2\nonumber\\
&+k\widehat\sigma\sum_i\|H^{1/2}d_i\|_2+4\widehat\sigma\sum_{i,v}\|e_{iv}\|_2,
\label{eq:objective}
\end{align}
where $\widehat\sigma$ is a residual scale and $k$ a penalty multiplier, both set below. The camera metric in~\eqref{eq:objective}, $\|H^{1/2}d_i\|_2=\sqrt{\sum_v\|M_vd_i\|_2^2}$, measures each 3D shift directly in projected image space and applies group sparsity~\cite{friedman2010group} across atoms and across individual observations. A coherent 3D shift $d$ costs $k\widehat\sigma\sqrt{\sum_v\|M_vd\|_2^2}$ through the physical term but $4\widehat\sigma\sum_v\|M_vd\|_2$ if explained as separate view errors. Since $\sqrt{\sum_v\|M_vd\|_2^2}\le\sum_v\|M_vd\|_2$ by Cauchy--Schwarz, any $k\le4$ makes the coherent explanation no more expensive than the view-by-view one, so cross-view shifts are assigned to $D$ while isolated outliers are routed to $E$.

\subsection{Candidate construction and optimization}
With reference coordinates, noise levels, and displaced atom identities withheld, candidate symmetries are inferred directly from the observations. We initialize correspondences via multistart species-constrained linear assignment in SciPy~\cite{virtanen2020scipy} and all-view least-squares triangulation. Spglib proposals across tolerances $\{0.005,0.01,0.02,0.05,0.10,0.20\}\,\angstrom$ generate candidate affine spaces; within each crystal system we keep the hypothesis with the most symmetry operations, and we add an unconstrained triclinic fallback (Supplementary Section~\ref{sec:implementation}).

For each candidate, we set $\widehat\sigma$ to the initial residual standard deviation (with a $0.001\,\angstrom$ floor) and profile out $\theta$ via orthogonal projection, leaving a convex problem in $(D,E)$. We solve it using FISTA~\cite{beck2009fista} and vector soft thresholding. Selected 3D and 2D supports are then jointly refitted with the affine base by unpenalized least squares to remove amplitude shrinkage.

\subsection{Withheld-view selection and output}
Five-fold cross-validation reconstructs from four views and predicts the fifth via species-constrained assignment under the radial Huber loss~\cite{huber1964robust} $\rho(r)=r^2/2$ for $r\leq4\widehat\sigma$ and $4\widehat\sigma(r-2\widehat\sigma)$ otherwise. Let $\ell_v(k)$ denote the loss on held-out view $v$ at multiplier $k$. For each $k\in\{2,3,4\}$, we first select the symmetry hypothesis with minimum $\sum_v\ell_v(k)$. Across $k$, letting $k_*$ minimize that sum and $\Delta_v(k)=\ell_v(k)-\ell_v(k_*)$, our primary paired one-SE rule~\cite{hastie2009elements} favors simpler departure supports by choosing the largest $k$ satisfying
\begin{equation}
\label{eq:one_se}
\overline\Delta(k)\leq \frac{\operatorname{sd}(\Delta_1(k),\ldots,\Delta_5(k))}{\sqrt5},
\end{equation}
where $\overline\Delta(k)$ is the mean of $\Delta_v(k)$ over the five folds. Alongside paired one-SE, we evaluate minimum-loss selection and a prespecified secondary fixed $k=4$; setting $D=0$ or $E=0$ yields component ablations. The final all-view refit outputs complete coordinates $X_G+D$, classified at $10^{-4}\,\angstrom$ so that retained departures are not masked by the symmetry-detection tolerance.

\subsection{Choosing localized or collective departures}
While the atomic penalty suits localized shifts, coordinated distortions are captured by constructing real isotypic displacement subspaces from each inferred parent symmetry, that is, displacement patterns that transform as irreducible representations of the parent group~\cite{campbell2006isodisplace}. Grouping equivalent irreducible components maintains basis invariance. Projecting out rigid translations and symmetry-invariant motions avoids redundancy with $X_G$. Normalizing each retained group basis $\widetilde U_{G,g}$ (of dimension $r_g$) through the cameras after profiling affine directions yields the collective branch, with mode amplitudes $c_g$:
\begin{equation}
\operatorname{vec}(D)=\sum_g\widetilde U_{G,g}c_g,\quad
\Omega_{\mathrm{mode}}=k\widehat\sigma\sum_g\sqrt{r_g/3}\,\|c_g\|_2.
\label{eq:mode}
\end{equation}
The factor $\sqrt{r_g/3}$ calibrates an $r_g$-dimensional subspace against a 3D single-atom shift; the error penalty and joint support refit are the same as in the atomic branch.

Both branches independently select their parent and $k$ via withheld-view validation. With $\gamma_v=\ell_v^{\mathrm{atom}}-\ell_v^{\mathrm{mode}}$ and $\overline\gamma$ its mean over folds, our adaptive selector switches to collective modes only when they achieve a consistent cross-view improvement under the same one-SE threshold as~\eqref{eq:one_se}:
\begin{equation}
\overline\gamma>\operatorname{sd}(\gamma_1,\ldots,\gamma_5)/\sqrt5.
\label{eq:switch}
\end{equation}
Selection sees only the observed views; the true parent group and the distortion family are never supplied (Supplementary Section~\ref{sec:modes} details mode construction and matched Huber refit controls).

\subsection{What changes when exact geometry is supplied?}
To evaluate perception and reasoning independently of reconstruction, let $R_{\mathrm{img}}(m)$ denote model $m$'s accuracy from rendered images and $R_{\mathrm{text}}(m)$ its accuracy when exact species, fractional coordinates, and lattice parameters replace the images as text. With $R_{\mathrm{geo}}$ the geometric reference defined above, we define the net gain $P$, residual deficit $S$, and closed deficit fraction $f$:
\begin{equation}
\label{eq:diagnostic}
\begin{aligned}
P&=R_{\mathrm{text}}-R_{\mathrm{img}},\\
S&=R_{\mathrm{geo}}-R_{\mathrm{text}},\\
f&=\frac{P}{R_{\mathrm{geo}}-R_{\mathrm{img}}}.
\end{aligned}
\end{equation}
Here $f$ measures how much of the visual error gap is closed by exact geometry, and $S$ captures residual failures in parsing, domain knowledge, and reasoning. Because $P$ is a net quantity, corrections and regressions cancel in it; tracking paired structure-level transitions separates the two.

\section{Experimental design}
\label{sec:setup}

To rigorously evaluate recoverability, coordinate fidelity, and downstream model response, our evaluations are organized into six standardized protocols across primary and external cohorts:

\textbf{Paired geometry substitution.} To measure sensitivity to supplied geometry, 17 vision-language models are evaluated on 210 primary Materials Project structures~\cite{jain2013commentary} (30 per crystal system) under both multi-view rendered images and exact text coordinates, using matched questions and three-answer majority voting. Task-supervised raster controls separately test usable image signal. ResNet-50~\cite{he2016resnet} and ViT small~\cite{dosovitskiy2021vit}, initialized from ImageNet~\cite{deng2009imagenet} weights, are fine-tuned with AdamW~\cite{loshchilov2019adamw} on 1,449 training structures and validated on 161 structures, with all five views of a structure kept together, then tested on the 210 primary structures by averaging logits across views. A 19-feature random forest~\cite{breiman2001forests} provides a numeric control trained on all 1,610 training structures. The non-frontier roster retains 14 of 15 attempted models under API-error and parsing gates at temperature 0.7, while the three frontier models use native defaults. Supplementary Sections~\ref{sec:benchmark} and~\ref{sec:ladder} specify the diagnostic benchmark, rendering, exact inputs, scoring, retries, refusals, and diagnostic controls.

\textbf{Geometric recoverability reference.} The recoverability reference evaluates multi-view 3D determination across 210 primary structures, a disjoint 1,950-structure census, and 210 replication structures. Five orthographic views supply complete species-labeled centroids and a known lattice. Because raster renderings display tiled cells with potential atomic overlaps, geometric recoverability and raster readability serve as distinct evaluation axes.

\textbf{Atomic reconstruction and preservation.} A reserved cohort of 28 parent structures (four per crystal system) comprising 1,568 test cases (originals, translations, and localized displacements under Gaussian noise or single-view corruption) evaluates whether reconstruction rejects measurement errors while preserving genuine physical departures. Every method receives identical complete centroids, species, cameras, and lattice. The pipeline was frozen prior to evaluation; identifier exclusions and structure matching strictly separate this cohort from the primary benchmark and historical development runs (see Supplementary Section~\ref{sec:provenance}).

\textbf{Evaluation endpoints and comparison baselines.} The primary atomic endpoint measures joint correctness of an original and its stable class-changing counterpart, alongside mean per-case RMSD after species-constrained matching to assess coordinate fidelity across all cases. Original-control accuracy measures performance on unchanged structures, while false restoration tracks whether a changed structure is incorrectly assigned its parent class. Huber triangulation serves as the external baseline. Retrospective no-refit comparisons and descriptive unadjusted 95\% parent-bootstrap intervals~\cite{efron1979bootstrap} complete the atomic evaluation.

\textbf{Transfer to collective departures.} Two disjoint MC3D PBE-v1 cohorts~\cite{huber2026mc3d} of 80 mode-eligible compositions each (960 cases per cohort) evaluate frozen pipelines across original, localized, and matched-norm collective distortion variants. Observations and perturbations are simulated from the relaxed MC3D structures. Atomic split, mode split, and adaptive atomic-or-mode split test representation choices against Huber refitting controls, including combined Huber. Cohort 1 prespecifies three coordinate criteria on collective cases, each requiring a minimum RMSD reduction and a multiplicity-adjusted material-bootstrap interval excluding zero: at least 15\% against atomic split under Gaussian noise, at least 15\% against atomic split under single-view corruption, and at least 5\% against combined Huber under corruption. Cohort 2 provides a secondary replication analysis (see Supplementary Section~\ref{sec:external}).

\textbf{Reconstructed-coordinate downstream handoff.} A 112-case follow-up reuses all 28 reserved atomic parents with original and large-displacement variants under frozen cameras, $0.01\,\angstrom$ Gaussian noise, and matched single-view corruption. Claude Sonnet 5.5 (Sonnet), Gemma 3 4B IT (Gemma), and Qwen3 4B receive identical lattice-only, Huber-coordinate, and split one-SE-coordinate prompts to predict crystal systems at a fixed $0.02\,\angstrom$ positional and $5^\circ$ angular tolerance; reference targets are computed at the same tolerances from each case's noise-free ground-truth structure, including its imposed displacement when present. Individual-run accuracies are summarized by mean and sample standard deviation alongside Sonnet's majority-vote analysis, providing a descriptive assessment of reconstructed-coordinate handoff.

\section{Results}
\label{sec:results}

\begin{table}[tbp]
\centering
\caption{\textbf{Exact geometry helps frontier models but leaves residual errors.} Each model uses 210 matched structures and complete reference geometry as text. Counts show cases corrected ($+$) or newly wrong ($-$) with text. Closed is $100(R_{\mathrm{text}}-R_{\mathrm{img}})/(1-R_{\mathrm{img}})$ relative to $R_{\mathrm{geo}}=1$, using unrounded counts. Frontier models use native API defaults; the non-frontier roster uses temperature 0.7. Claude Fable 5.1 has four text cases with only refusals, counted incorrect. Correction opportunities vary with image-baseline error.}
\label{tab:vlm_substitution}
\footnotesize
\setlength{\tabcolsep}{2pt}
\renewcommand{\arraystretch}{1.08}
\begin{tabular*}{\columnwidth}{@{\extracolsep{\fill}}lrrrrr@{}}
\toprule
&\multicolumn{2}{c}{Accuracy (\%)}&\multicolumn{2}{c}{Paired counts}&Closed\\
\cmidrule(lr){2-3}\cmidrule(lr){4-5}
Model & Image $\uparrow$ & Text $\uparrow$ & $+$ & $-$ & (\%) $\uparrow$\\
\midrule
\multicolumn{6}{c}{\textbf{Frontier}}\\
\midrule
Claude Fable 5.1 & 74.8 & 91.0 & 44 & 10 & 64.2\\
Claude Sonnet 5.5 & 72.4 & 91.9 & 54 & 13 & 70.7\\
Claude Opus 5.5 & 69.0 & 79.0 & 45 & 24 & 32.3\\
\midrule
Gemini 3.6 Flash & 73.3 & 85.2 & 45 & 20 & 44.6\\
Grok 4.5 & 61.4 & 85.2 & 69 & 19 & 61.7\\
Claude Opus 4.8 & 58.1 & 66.7 & 62 & 44 & 20.5\\
Llama 4 Maverick & 44.3 & 49.5 & 63 & 52 & 9.4\\
GLM-4.6V & 44.3 & 44.8 & 63 & 62 & 0.9\\
Qwen3-VL 8B & 37.6 & 45.2 & 71 & 55 & 12.2\\
GPT-4.1 mini & 36.7 & 41.4 & 60 & 50 & 7.5\\
Qwen3-VL 235B & 33.3 & 54.3 & 73 & 29 & 31.4\\
Seed 1.6 & 25.7 & 42.4 & 76 & 41 & 22.4\\
Mistral Medium 3.1 & 22.9 & 45.2 & 72 & 25 & 29.0\\
Qwen3-VL 32B & 22.9 & 45.2 & 66 & 19 & 29.0\\
Llama 4 Scout & 20.5 & 50.5 & 89 & 26 & 37.7\\
Nova Pro & 18.1 & 46.7 & 73 & 13 & 34.9\\
Mistral Small 2603 & 14.8 & 45.2 & 84 & 20 & 35.8\\
\bottomrule
\end{tabular*}
\end{table}

\begin{table}[t]
\centering
\caption{\textbf{Split reconstruction across selection rules and component ablations.} The reserved cohort has 28 parents, 1,120 Gaussian cases and 448 corrupted cases. RMSD is mean per-case error in m\AA. Joint correctness requires both original and changed labels to be correct; false restoration assigns the changed case its parent class. Joint/restoration denominators are 420/184 and original-control denominators 280/224 (Gaussian/corrupted). One-SE is development-selected primary; fixed $k=4$ is prespecified secondary. Component-removal rows retain their recorded selection rules. $^\dagger$The retrospective no-refit ablation includes changes in fold losses and selection. Bold identifies full-method variants.}
\label{tab:reconstruction}
\footnotesize
\setlength{\tabcolsep}{1.5pt}
\renewcommand{\arraystretch}{0.95}

\begin{tabular*}{\columnwidth}{@{\extracolsep{\fill}}>{\raggedright\arraybackslash}p{0.36\columnwidth}rrrr@{}}
\toprule
\multicolumn{5}{c}{\textbf{Gaussian noise}}\\
\midrule
Configuration & RMSD $\downarrow$ & \shortstack{Joint $\uparrow$\\(\%)} & \shortstack{Restored $\downarrow$\\(\%)} & \shortstack{Original $\uparrow$\\(\%)}\\
\midrule
Huber & 5.271 & 75.7 & 9.5 & 92.9\\
\addlinespace[1pt]
No physical shifts ($D=0$) & 3.921 & 81.7 & 11.0 & 96.4\\
No view errors ($E=0$) & 3.377 & 83.3 & 0.0 & 87.9\\
One-SE, no refit$^\dagger$ & 3.486 & 76.0 & 0.0 & 82.1\\
\addlinespace[1pt]
\textbf{Split, min-loss} & 3.387 & 83.3 & 0.0 & 87.9\\
\textbf{Split, one-SE} & 3.336 & 86.2 & 0.2 & 90.7\\
\textbf{Split, $k=4$} & 3.325 & 88.3 & 0.7 & 93.6\\
\midrule
\multicolumn{5}{c}{\textbf{One corrupted observation}}\\
\midrule
Configuration & RMSD $\downarrow$ & \shortstack{Joint $\uparrow$\\(\%)} & \shortstack{Restored $\downarrow$\\(\%)} & \shortstack{Original $\uparrow$\\(\%)}\\
\midrule
Huber & 7.395 & 85.9 & 0.0 & 88.8\\
\addlinespace[1pt]
No physical shifts ($D=0$) & 5.026 & 93.5 & 0.0 & 96.4\\
No view errors ($E=0$) & 8.757 & 37.0 & 0.0 & 46.4\\
One-SE, no refit$^\dagger$ & 5.888 & 90.8 & 0.0 & 92.4\\
\addlinespace[1pt]
\textbf{Split, min-loss} & 4.421 & 85.9 & 0.0 & 88.8\\
\textbf{Split, one-SE} & 4.349 & 90.8 & 0.0 & 92.9\\
\textbf{Split, $k=4$} & 4.348 & 93.5 & 0.0 & 95.5\\
\bottomrule
\end{tabular*}
\end{table}

\subsection{Frontier models remain sensitive to supplied geometry}
On the 210 primary structures, the three tested frontier models misclassify 53--65 image-input cases (Table~\ref{tab:vlm_substitution}). Bypassing visual extraction by supplying exact 3D reference geometry as text raises accuracy by 10.0--19.5 percentage points, yet leaves 17--44 cases incorrect. With exact geometry, each frontier model corrects 44--54 image-input errors and introduces 10--24 new errors (Claude Fable 5.1 (Fable) has four text cases with only refusals).

All 14 non-frontier models also improve in aggregate, closing a median 29.0\% of their deficit to the primary geometric reference $R_{\mathrm{geo}}=1$. API settings held fixed across input formats within each model. Moreover, a correct symmetry label leaves coordinate fidelity unresolved: under the benchmark's metric-based rule, lattice parameters alone suffice for 140 of the 210 primary cases without inspecting atomic sites. Auditing structural understanding thus requires evaluating the 3D intermediate directly. First establishing multi-view recoverability, and then testing whether reconstruction from imperfect views preserves physical departures.

\begin{figure*}[t]
\centering
\includegraphics[width=0.85\textwidth]{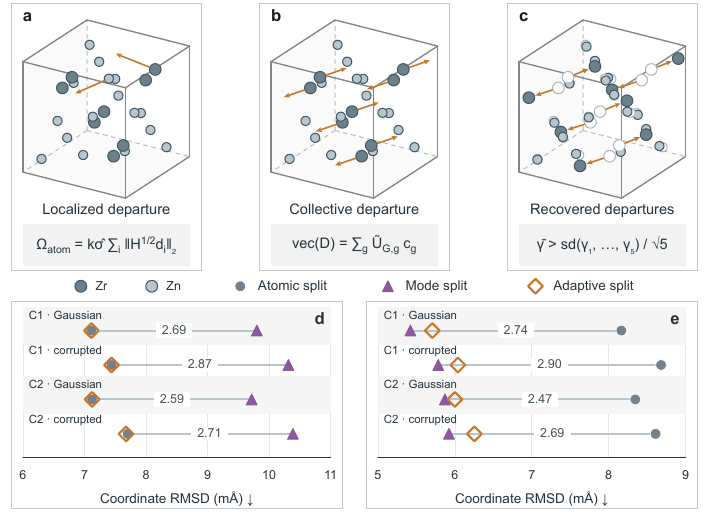}
\caption{\textbf{Adaptive reconstruction balances localized and collective recovery.} Reference two-site (a) and collective symmetry-mode (b) perturbations of ZrZn$_2$ with equal displacement norm after translation removal. (c) Adaptive split reconstructs the collective field from five Gaussian-noisy views on a lower-median-error Cohort 1 example (RMSD 5.3 m\AA; cells share viewpoint and scale; departures magnified 36×; hollow circles mark parent sites and filled circles mark magnified recovered positions in (c)). Mean coordinate RMSD for localized (d) and collective (e) departures across both cohorts and conditions (160 cases from 80 materials per marker; lower is better). Gray lines connect method means with endpoint gaps in m\AA; hollow vermilion diamonds show adaptive split tracking the stronger representation.}
\label{fig:collective}
\end{figure*}

\subsection{Recoverability provides an independent geometric reference}
Our model-free render ceiling shows that projective ambiguity vanishes as calibrated views are added. Across the 1,950-structure census (threshold $10^{-6}\,\angstrom$), projective phantoms occur in 15.4\% of two-view and 1.9\% of three-view combinations, but in 0\% of four- or five-view combinations. In all 9,750 four-view cases, each site had at least one view in which its centroid was distinct from every other same-species centroid, so removing it would change the observations. Using five views, the finite-tolerance reference recovers labels with 100\% accuracy on the primary structures and census, and 99.5\% on the replication structures at $\delta=0.01\,\angstrom$ (95.2\%, 87.8\%, and 91.0\% at the $0.15\,\angstrom$ tolerance used in the released benchmark).

While the render ceiling characterizes complete geometric observations, task-supervised raster controls separately confirm usable image signal: ResNet-50 and ViT-small reach 89.5\% and 83.3\% accuracy, respectively. Because neither label prediction nor supervised raster classification certifies 3D coordinate extraction, we next evaluate how faithfully split reconstruction recovers coordinates from imperfect centroids.

\subsection{Split reconstruction improves coordinates across selection rules}
Symmetry-aware split reconstruction consistently improves 3D coordinate fidelity across all three full-method selection variants (Table~\ref{tab:reconstruction}). On the 1,568 reserved cases, paired one-SE reduces coordinate RMSD by 36.7\% under Gaussian noise and 41.2\% under corruption relative to Huber triangulation. Against the symmetry-constrained fit with physical shifts disabled ($D=0$), the reductions are 14.9\% and 13.5\%, respectively (Supplementary Section~\ref{sec:atomic_results}).

\textbf{Preventing false restorations and outlier absorption.} Component ablations confirm that the 3D physical ($D$) and 2D error ($E$) branches prevent distinct failure modes. Under Gaussian noise, removing physical departures ($D=0$) forces symmetry-broken structures onto parent space groups, causing 46 false restorations among 420 changed cases versus just one for full one-SE reconstruction. Conversely, removing the observation-error branch ($E=0$) absorbs single-view outliers into 3D coordinates, raising corrupted-view RMSD to 8.757 m\AA\ and lowering joint correctness to 37.0\%, compared to 4.349 m\AA\ and 90.8\% for full one-SE (under pure Gaussian noise, its 1.2\% reduction against $E=0$ has an interval spanning zero).

\textbf{Joint support refitting.} Retaining penalized physical and error coefficients across folds and final estimation raises one-SE RMSD to 3.486 m\AA\ under Gaussian noise and 5.888 m\AA\ under corruption (Supplementary Section~\ref{sec:refit}).

\textbf{Selection and structural preservation.} All full-method variants correctly classify 100\% of the 56 noise-free originals and agree on 99.6\% of the 280 original/translation pairs. Under noise and corruption, fixed $k=4$ yields higher joint correctness than one-SE and higher original-control accuracy, while producing three Gaussian false restorations rather than one. Paired one-SE retains 96.7\% (406/420, Gaussian) and 99.5\% (183/184, corrupted) of stable changed labels, balancing distortion recovery against parent preservation.

\subsection{Adaptive representations extend recovery to collective departures}
Figure~\ref{fig:collective}a--c shows reference localized and collective perturbations of ZrZn$_2$ and the collective field recovered by adaptive split; Figure~\ref{fig:collective}d--e compares the three representations. Adaptive atomic-or-mode split retains localized accuracy while capturing the collective benefit.

Adaptive split reduces collective RMSD against atomic split by 30.1\% and 30.4\% in Cohort 1 and by 28.1\% and 27.4\% in Cohort 2 under Gaussian noise and corruption, respectively, and outperforms combined Huber under corruption by 6.6\% and 6.1\% (satisfying all three prespecified Cohort 1 criteria while matching combined Huber under Gaussian noise).

Mode-only fits remain slightly tighter on pure collective cases but incur high localized error. Across 1,280 original/localized controls, adaptive split correctly classifies 1,253 cases versus 1,255 for atomic split, with six regressions offset by four corrections and at most a 1.7\% mean coordinate-error increase in any control cell (Supplementary Section~\ref{sec:external_controls}).

\begin{table}[t]
\centering
\caption{\textbf{Split coordinates improve downstream accuracy.} Accuracy (\%), mean $\pm$ sample SD over three runs on 112 cases from 28 parents. Malformed or truncated outputs count as incorrect. Split denotes one-SE reconstruction. Bold and underline indicate the highest and second-highest mean accuracy in each row.}
\label{tab:coordinate_handoff}
\footnotesize
\setlength{\tabcolsep}{2pt}
\renewcommand{\arraystretch}{1.1}
\begin{tabular*}{\columnwidth}{@{\extracolsep{\fill}}lccc@{}}
\toprule
Model & Lattice only & Huber & Split (ours)\\
\midrule
Sonnet 5.5 & $42.26 \pm 1.03$ & $\underline{65.18} \pm 2.36$ & $\mathbf{75.00} \pm 1.79$\\
Gemma 3 4B IT & $\underline{8.33} \pm 1.03$ & $7.74 \pm 1.03$ & $\mathbf{8.63} \pm 1.36$\\
Qwen3 4B & $\mathbf{16.67} \pm 2.73$ & $13.99 \pm 2.25$ & $\underline{15.48} \pm 2.87$\\
\bottomrule
\end{tabular*}
\end{table}

\subsection{Recovered coordinates improve the matched Huber comparison}
Handing reconstructed coordinates from the 112 noisy and corrupted follow-up cases to downstream models improves mean classification accuracy over Huber coordinates across all three tested models (Table~\ref{tab:coordinate_handoff}; Supplementary Section~\ref{sec:handoff}). Sonnet achieves the largest gain, improving by 9.82 percentage points, while Gemma and Qwen3 4B improve by 0.89 and 1.49 points, respectively. Under three-answer majority voting, Sonnet achieves 75.0\% with split coordinates versus 67.9\% with Huber and 42.9\% with lattice alone.

Comparing frontier and local models highlights how geometric recovery interacts with reasoning capacity. While Sonnet translates denoised 3D coordinates into a gain over Huber, the smaller 4B models remain bottlenecked by textual coordinate parsing, scoring below 17\% across all inputs. Qwen3 4B correctly classifies none of the 46 stable class-changing cases from either Huber or split coordinates in any of the three runs. Faithful 3D geometric recovery thus supplies the critical structural evidence whose downstream utility scales with model reasoning capacity.
\section{Conclusion and limitations}
\label{sec:discussion}

Evaluating structural understanding in vision-language models requires decoupling visual perception from downstream reasoning through an explicit, inspectable 3D geometric intermediate. Our model-free \emph{render ceiling} establishes exact 3D site determination from calibrated projections by ruling out projective phantoms, showing that four to five orthographic views uniquely determine crystal geometry across a 1,950-structure census. Against this reference, our paired coordinate-substitution diagnostic across 17 VLMs shows that frontier models make substantial errors on multi-view crystal renderings, and that supplying exact 3D geometry closes up to 70.7\% of the accuracy gap while exposing persistent reasoning errors and regressions. Under measurement noise and single-view corruption, our symmetry-aware split reconstruction reduces coordinate error over Huber triangulation while preserving genuine symmetry-breaking displacements. Withheld-view adaptive selection unites localized atomic shifts and collective distortion modes, cutting collective coordinate error across two external cohorts while preserving localized fidelity. Handing the reconstructed coordinates to downstream models improves classification accuracy over Huber-triangulated inputs across all tested models.

Our guarantees operate within defined geometric boundaries. Theorem~\ref{thm:exact} guarantees exact recovery at zero tolerance when projective phantoms are absent; the finite-tolerance implementation $R_{\mathrm{geo}}(\delta)$ is an empirical reference rather than a universal classifier upper bound, and the FISTA convergence check certifies the fixed-candidate convex subproblem rather than global correspondence optimality or formal support recovery. The reconstruction pipeline assumes calibrated orthographic cameras, a known lattice, and complete, species-labeled 2D centroids; missing detections, centroid extraction from raw raster images, camera-pose uncertainty, and experimental electron-microscopy image formation remain directions for end-to-end integration. The external collective-mode evaluations use simulated perturbations of mode-eligible compositions rather than DFT-relaxed defect trajectories, and the 112-case handoff is a descriptive within-cohort comparison in which 4B local models remain bottlenecked by textual coordinate parsing and Sonnet's majority-vote bootstrap interval spans zero. Pairing automated 2D atomic-centroid detectors with split reconstruction offers a direct path toward fully verifiable, end-to-end scientific vision pipelines.
\section*{Data and code availability}
\label{sec:availability}
Code, structure and split manifests, evaluation schedules, prompts, exact-geometry inputs, and predictions are available at \url{https://github.com/KurbanIntelligenceLab/render-ceiling}. The included reproducibility guide documents the Python API and command-line interface, and the results index provides record checksums.

\enlargethispage{\baselineskip}
\section*{LLM Usage Statement}
\begingroup\small
We used five LLM-based tools. Claude Opus 5.5 (Opus; Anthropic) assisted with literature search, code, and manuscript preparation. Claude Fable 5.1 (Anthropic) assisted with literature search and manuscript preparation. Gemini 4 Argon (Google) assisted with manuscript preparation. GPT-6 Astra (OpenAI) assisted with code, experiments, numerical verification of reported values, plotting code for Figures 1--2, and manuscript preparation. GPT-6.1 Sol (OpenAI) assisted with code. The authors developed the method; the coding assistants helped optimize its implementation and analyze experimental results. For selected implementation tasks, coding tools autonomously inspected repository files, wrote and executed code, and verified outputs in response to author-defined requests, with authors reviewing the resulting changes. For the manuscript, the tools drafted and edited text, which the authors then revised in full. Opus and Fable also appear as evaluated models in Table 1; their benchmark outputs were collected only through the evaluation pipeline described in Section 4, independently of their use as research assistants. The authors take full responsibility for all claims, references, and interpretation.
\par\endgroup

\clearpage
\bibliography{references}
\clearpage
\setcounter{section}{0}
\setcounter{table}{0}
\setcounter{figure}{0}
\setcounter{equation}{0}
\renewcommand{\thesection}{S\arabic{section}}
\renewcommand{\thetable}{S\arabic{table}}
\renewcommand{\thefigure}{S\arabic{figure}}
\renewcommand{\theequation}{S\arabic{equation}}
\renewcommand{\theHsection}{supp.\arabic{section}}
\renewcommand{\theHtable}{supp.\arabic{table}}
\renewcommand{\theHfigure}{supp.\arabic{figure}}
\renewcommand{\theHequation}{supp.\arabic{equation}}
\section*{Supplementary Material}
\section{Reconstruction implementation}
\label{sec:implementation}
\paragraph{Correspondence and symmetry.} Multistart species-constrained assignment alternates with all-view least-squares triangulation. With true noise withheld, the initial scale is set to $\widehat\sigma=\sqrt{\max(\mathrm{SSE}/[N(2V-3)],10^{-6})}$. For each proposed symmetry, we average the Cartesian actions of periodic operations and species-preserving atom permutations, take the fixed subspace, and add three global translations. Projection uses the camera metric $I_N\otimes H$. Proposals whose projected reference fails the crystal-system check at $10^{-4}\,\angstrom$ are discarded; the triclinic fallback remains unconstrained.

\paragraph{Optimization and support.} Stacking cameras as $A=[M_1;\ldots;M_V]$ with $\mathcal A=I_N\otimes A$, let $U$ be an orthonormal basis for the columns of $\mathcal A B_G$. The projector $Q=I-UU^{\mathsf T}$ removes affine nuisance directions, while whitened coefficients $u_i=H^{1/2}d_i$ have forward map $I_N\otimes(AH^{-1/2})$. We project the target and physical/error designs by $Q$ prior to group soft thresholding. Accepted fits satisfy a scaled proximal-residual threshold of $10^{-5}$. Groups with coefficient norm exceeding $10^{-5}\max(\widehat\sigma,0.001)$ enter the joint unpenalized refit (relative least-squares cutoff $10^{-9}$).

\paragraph{Selection and comparators.} Correspondence, scale, and candidate spaces are rebuilt inside each withheld-view fold; valid hypotheses must exist across all folds and the full fit. The null-guard ablation defaults to observation-only ($D=0$) unless its mean excess fold loss over paired one-SE exceeds one paired standard error. Huber triangulation uses a leverage-adjusted median-absolute-deviation scale ($0.001\,\angstrom$ floor), radial threshold $4\widehat\sigma$, and classification tolerance $\max(0.005,6u)\,\angstrom$ (where $u$ is the maximum coordinate-uncertainty scale), without symmetry denoising.

\section{Collective modes and matched controls}
\label{sec:modes}
\paragraph{Mode construction.} Symmetry operations act via Cartesian rotations and species-preserving permutations. Jointly diagonalizing symmetric sums over conjugacy classes yields real isotypic displacement subspaces (including equivalent and conjugate components), from which translations, symmetry-invariant motions, and projected null directions are removed. For each retained basis $U_{G,g}$ with fixed numerical rank, camera normalization after nuisance projection is given by
\begin{equation}
\widetilde U_{G,g}=U_{G,g}\left[(Q\mathcal A U_{G,g})^{\mathsf T}(Q\mathcal A U_{G,g})\right]^{-1/2},
\label{eq:whiten}
\end{equation}
which makes the retained projected columns orthonormal so basis rotations within a group preserve both its norm and prediction space.

\paragraph{Matched refitting.} Mode Huber and mode split share the profiled support-selection objective: mode Huber refits physical coordinates under radial Huber loss, whereas mode split jointly refits the selected physical and observation-error supports. Combined Huber applies the adaptive atomic-or-mode selector to the mode Huber branch. Both pipelines infer candidate parents solely from observations and compare their mode branch against atomic split via withheld-view prediction.

\section{Original diagnostic benchmark}
\label{sec:benchmark}
To prevent split leakage, evaluation structures contain at least one of 13 reserved elements excluded from the training split. Five fixed orthographic cameras render atomic discs, species colors, and cell edges at $768\times768$ pixels. Raster images tile the conventional cell $2\times2\times2$, whereas the geometric reference uses a single conventional cell with all centroids retained.

\paragraph{Supervised controls.} Task-supervised ResNet-50~\cite{he2016resnet} and ViT-small~\cite{dosovitskiy2021vit} are trained on 1,449 structures and validated on 161 (all views of a structure in one partition), then tested on the 210 primary structures. Initialized from ImageNet~\cite{deng2009imagenet} weights on $224\times224$ normalized crops without augmentation, both train for 8 epochs with AdamW~\cite{loshchilov2019adamw} (batch size 64, learning rate $3\times10^{-4}$, weight decay 0.01, cosine annealing, mixed precision) and average logits across the five views at inference. A 19-feature random forest~\cite{breiman2001forests} numeric control is trained on all 1,610 training structures.

\section{Paired-model protocol and diagnostic controls}
\label{sec:ladder}
\paragraph{Inputs and scoring.} Each model receives five PNG views or exact text (six lattice parameters plus species and fractional coordinates per atom) followed by an identical seven-way crystal-system prompt, scored by three-repeat majority vote (ties broken by repeat order; last valid label parsed). Of 15 attempted non-frontier models (evaluated via OpenRouter at temperature 0.7), 14 satisfy the 5\% API-error and parsing gates.

\paragraph{Frontier extension.} Claude Fable 5.1 (Fable), Claude Sonnet 5.5 (Sonnet), and Claude Opus 5.5 (Opus) are evaluated via the native Anthropic Messages API under provider defaults with prompt caching. Initial output caps of 4,096 tokens (including reasoning) are expanded to 16,384 tokens (and 32,768 for Fable) only upon truncation; completed responses are never retried on correctness. All 1,260 evaluation slots per model completed without transport failure or unresolved truncation. Sonnet and Opus yield three valid votes per case; Fable produced 12 text refusals across four cases, counted as incorrect where no valid vote remained (retaining the 210-case denominator).

\begin{table*}[t]
\centering
\caption{\textbf{Complete paired outcomes for image and exact-geometry inputs.} Each model has 210 matched structures. Corrected means wrong with images and correct with text; newly wrong means the reverse. Text accuracy was higher for all 17 models, with 11 gains significant at $\alpha=0.05$ after Holm adjustment of two-sided exact McNemar tests.}
\label{tab:vlm_paired_counts}
\small
\setlength{\tabcolsep}{4pt}
\renewcommand{\arraystretch}{1.08}
\begin{tabular*}{\textwidth}{@{\extracolsep{\fill}}lrrrrrrr@{}}
\toprule
Model & \shortstack{Image\\correct} & \shortstack{Text\\correct} & \shortstack{Both\\correct} & Corrected & \shortstack{Newly\\wrong} & \shortstack{Both\\wrong} & \shortstack{Closed\\(\%)}\\
\midrule
\multicolumn{8}{c}{\textbf{Frontier}}\\
\midrule
Claude Fable 5.1 & 157/210 & 191/210 & 147 & 44 & 10 & 9 & 64.2\\
Claude Sonnet 5.5 & 152/210 & 193/210 & 139 & 54 & 13 & 4 & 70.7\\
Claude Opus 5.5 & 145/210 & 166/210 & 121 & 45 & 24 & 20 & 32.3\\
\midrule
Gemini 3.6 Flash & 154/210 & 179/210 & 134 & 45 & 20 & 11 & 44.6\\
Grok 4.5 & 129/210 & 179/210 & 110 & 69 & 19 & 12 & 61.7\\
Claude Opus 4.8 & 122/210 & 140/210 & 78 & 62 & 44 & 26 & 20.5\\
Llama 4 Maverick & 93/210 & 104/210 & 41 & 63 & 52 & 54 & 9.4\\
GLM-4.6V & 93/210 & 94/210 & 31 & 63 & 62 & 54 & 0.9\\
Qwen3-VL 8B & 79/210 & 95/210 & 24 & 71 & 55 & 60 & 12.2\\
GPT-4.1 mini & 77/210 & 87/210 & 27 & 60 & 50 & 73 & 7.5\\
Qwen3-VL 235B & 70/210 & 114/210 & 41 & 73 & 29 & 67 & 31.4\\
Seed 1.6 & 54/210 & 89/210 & 13 & 76 & 41 & 80 & 22.4\\
Mistral Medium 3.1 & 48/210 & 95/210 & 23 & 72 & 25 & 90 & 29.0\\
Qwen3-VL 32B & 48/210 & 95/210 & 29 & 66 & 19 & 96 & 29.0\\
Llama 4 Scout & 43/210 & 106/210 & 17 & 89 & 26 & 78 & 37.7\\
Nova Pro & 38/210 & 98/210 & 25 & 73 & 13 & 99 & 34.9\\
Mistral Small 2603 & 31/210 & 95/210 & 11 & 84 & 20 & 95 & 35.8\\
\bottomrule
\end{tabular*}
\end{table*}

\paragraph{Diagnostic control findings.} Lattice metrics alone determine 140/210 primary labels (60 of the remaining 70 involve hexagonal/trigonal ambiguities). The numeric random forest reaches 188/210 on crystal systems (89.5\%) versus 143/210 on space groups across 44 represented groups; it matches ResNet-50's 188/210 in count but not in cases. A separate diagnostic pixel-to-coordinate baseline yields 30/210 downstream accuracy compared to 106/210 when supplying exact geometry to the same model.

\section{Evaluation provenance and definitions}
\label{sec:provenance}
\paragraph{Development and cohort isolation.} Selector refinement used a 14-parent development set and an initial 28-parent exploration cohort, where paired one-SE was selected by maximizing the worse of Gaussian/corrupted joint correctness (followed by their sum and simplicity tie-breaking). The primary evaluation uses a strictly disjoint second reserved cohort of 28 parents under frozen rules, excluding all primary benchmark identifiers and 84 earlier method-study parents (verified via primitive-reduced structure matching at lattice tolerance $10^{-4}$, site tolerance $10^{-3}$, and angle tolerance $0.01^\circ$).

\paragraph{Atomic cohort and metrics.} The 28 reserved parents (four per crystal system; 3--32 atoms; minimum separation $\ge 0.7\,\angstrom$; stable symmetry across $0.005, 0.01, 0.02\,\angstrom$) generate unmodified originals, translated controls, and two-atom displaced variants ($0.03$ or $0.10\,\angstrom$). Across two five-camera sets, Gaussian noise ($\sigma\in\{0,0.003,0.01\}\,\angstrom$) yields 1,120 cases, and a single-view $0.10\,\angstrom$ centroid outlier yields 448 corrupted cases (1,568 total). Reference labels are scored at $0.01\,\angstrom$, while reconstructed outputs are classified at $10^{-4}\,\angstrom$ as in the main text; stable-change endpoints require consistent reference changes across all three tolerances (420 Gaussian and 184 corrupted pairs). Coordinate RMSD is averaged per case after species-constrained Cartesian matching without rigid alignment. Unadjusted 95\% parent-bootstrap intervals~\cite{efron1979bootstrap} use 5,000 draws resampling all descendant cases together.

\section{External sampling and analysis provenance}
\label{sec:external}
\paragraph{Cohort construction and protocol.} Two disjoint external cohorts of 80 reduced compositions each are sampled from a 2,000-entry MC3D frame (excluding local compositions). Eligible parents contain 3--32 sites (separation $\ge 0.7\,\angstrom$), stable symmetry across $\{0.0001, 0.005, 0.01, 0.02\}\,\angstrom$, and a nontrivial collective displacement group below half the Cartesian dimension engaging at least three atoms. Each parent provides an original, a mean-centered two-site localized perturbation, and a norm-matched collective mode perturbation across two five-camera sets under $0.01\,\angstrom$ Gaussian noise with or without a $0.10\,\angstrom$ single-view outlier (960 cases per cohort: 320 collective and 640 original/localized controls).

\paragraph{Analysis roles.} Cohort 1 prospectively evaluates the frozen adaptive method on three prespecified collective-coordinate comparisons---against atomic split under Gaussian noise and under corruption (each requiring $\ge15\%$ RMSD reduction) and against combined Huber under corruption ($\ge5\%$)---with Bonferroni-adjusted 98.333\% parent-bootstrap intervals (20,000 draws) required to exclude zero, alongside control-preservation gates. Cohort 2 is a secondary replication of the unchanged pipeline. The structure in Figure~2c of the main text is the unfiltered lower-median RMSD case among the 80 Cohort-1 Gaussian cases (arrows $<2$~pt omitted).

\section{Detailed reconstruction results}
\label{sec:atomic_results}
\begin{table}[t]
\centering
\caption{\textbf{Coordinate intervals distinguish clear gains from close comparisons.} Relative RMSD reductions of the primary paired one-SE model on the reserved atomic cohort. Intervals are descriptive 95\% parent-bootstrap intervals from 5,000 draws and are unadjusted for multiplicity. Intervals crossing zero do not establish a reduction.}
\label{tab:atomic_intervals}
\small\setlength{\tabcolsep}{3pt}
\begin{tabular}{@{}lrr@{}}
\toprule
Comparator & Reduction (\%) & 95\% interval\\
\midrule
\multicolumn{3}{@{}l}{\textit{Gaussian}}\\
Huber triangulation & 36.7 & $[29.2, 43.9]$\\
Observation-only & 14.9 & $[10.0, 20.5]$\\
Physical-only & 1.2 & $[-0.9, 3.9]$\\
Split, minimum loss & 1.5 & $[-0.5, 4.2]$\\
Split, fixed $k=4$ & -0.3 & $[-2.1, 1.1]$\\
Split, null guard & 1.2 & $[-1.1, 3.9]$\\
\addlinespace[3pt]
\multicolumn{3}{@{}l}{\textit{One corrupted observation}}\\
Huber triangulation & 41.2 & $[34.1, 48.3]$\\
Observation-only & 13.5 & $[7.8, 20.0]$\\
Physical-only & 50.3 & $[45.8, 55.3]$\\
Split, minimum loss & 1.6 & $[0.3, 3.3]$\\
Split, fixed $k=4$ & 0.0 & $[-0.6, 0.5]$\\
Split, null guard & 0.4 & $[-0.3, 1.3]$\\
\bottomrule
\end{tabular}
\end{table}

\begin{table}[t]
\centering
\caption{\textbf{Gaussian reconstruction balances coordinate recovery and change preservation.} Second reserved cohort, 1,120 cases; RMSD is in $10^{-3}\,\angstrom$.}
\label{tab:atomic_gaussian}
\small\setlength{\tabcolsep}{2pt}
\begin{tabular}{@{}rrrrrr@{}}
\toprule
Overall & Originals & \shortstack{Stable\\changes} & Joint & Restored & RMSD\\
\midrule
\multicolumn{6}{@{}l}{\textit{Huber triangulation}}\\
960/1120 & 260/280 & 339/420 & 318/420 & 40/420 & 5.271\\
\addlinespace[2pt]
\multicolumn{6}{@{}l}{\textit{Observation-only}}\\
1026/1120 & 270/280 & 363/420 & 343/420 & 46/420 & 3.921\\
\addlinespace[2pt]
\multicolumn{6}{@{}l}{\textit{Physical-only}}\\
1036/1120 & 246/280 & 408/420 & 350/420 & 0/420 & 3.377\\
\addlinespace[2pt]
\multicolumn{6}{@{}l}{\textit{Split, minimum loss}}\\
1036/1120 & 246/280 & 408/420 & 350/420 & 0/420 & 3.387\\
\addlinespace[2pt]
\multicolumn{6}{@{}l}{\textit{Split, paired one-SE}}\\
1050/1120 & 254/280 & 406/420 & 362/420 & 1/420 & 3.336\\
\addlinespace[2pt]
\multicolumn{6}{@{}l}{\textit{Split, fixed $k=4$}}\\
1064/1120 & 262/280 & 403/420 & 371/420 & 3/420 & 3.325\\
\addlinespace[2pt]
\multicolumn{6}{@{}l}{\textit{Split, null guard}}\\
1058/1120 & 264/280 & 395/420 & 367/420 & 12/420 & 3.378\\
\bottomrule
\end{tabular}
\end{table}

\begin{table}[t]
\centering
\caption{\textbf{Observation-error modeling preserves reconstruction under a corrupted centroid.} Second reserved cohort, 448 cases; RMSD is in $10^{-3}\,\angstrom$.}
\label{tab:atomic_corrupted}
\small\setlength{\tabcolsep}{2pt}
\begin{tabular}{@{}rrrrrr@{}}
\toprule
Overall & Originals & \shortstack{Stable\\changes} & Joint & Restored & RMSD\\
\midrule
\multicolumn{6}{@{}l}{\textit{Huber triangulation}}\\
405/448 & 199/224 & 173/184 & 158/184 & 0/184 & 7.395\\
\addlinespace[2pt]
\multicolumn{6}{@{}l}{\textit{Observation-only}}\\
432/448 & 216/224 & 180/184 & 172/184 & 0/184 & 5.026\\
\addlinespace[2pt]
\multicolumn{6}{@{}l}{\textit{Physical-only}}\\
325/448 & 104/224 & 183/184 & 68/184 & 0/184 & 8.757\\
\addlinespace[2pt]
\multicolumn{6}{@{}l}{\textit{Split, minimum loss}}\\
422/448 & 199/224 & 183/184 & 158/184 & 0/184 & 4.421\\
\addlinespace[2pt]
\multicolumn{6}{@{}l}{\textit{Split, paired one-SE}}\\
431/448 & 208/224 & 183/184 & 167/184 & 0/184 & 4.349\\
\addlinespace[2pt]
\multicolumn{6}{@{}l}{\textit{Split, fixed $k=4$}}\\
436/448 & 214/224 & 182/184 & 172/184 & 0/184 & 4.348\\
\addlinespace[2pt]
\multicolumn{6}{@{}l}{\textit{Split, null guard}}\\
432/448 & 210/224 & 183/184 & 169/184 & 0/184 & 4.366\\
\bottomrule
\end{tabular}
\end{table}

Paired one-SE improves corrupted-view joint correctness over minimum-loss selection by 4.89 percentage points (95\% parent-bootstrap interval $[2.17,8.15]$), while differing from observation-only ($D=0$) by $+4.52$ points $[-2.63,10.44]$ under Gaussian noise and $-2.72$ points $[-5.43,0.54]$ under corruption. Fixed $k=4$ achieves slightly higher joint counts at the cost of three Gaussian false restorations (versus one for one-SE), whereas the null-guard variant falsely restored 12 changed structures and was not adopted.

\section{Support-refit ablation}
\label{sec:refit}
\begin{table}[t]
\centering
\caption{\textbf{Joint support refitting improves atomic coordinate recovery.} The ablation reuses the second reserved cohort. RMSD is in $10^{-3}\,\angstrom$.}
\label{tab:refit}
\small\setlength{\tabcolsep}{3pt}
\begin{tabular}{@{}lrrr@{}}
\toprule
Method & RMSD & Joint & Restored\\
\midrule
\multicolumn{4}{@{}l}{\textit{Gaussian}}\\
Paired one-SE, no refit & 3.486 & 319/420 & 0/420\\
Paired one-SE, refit & 3.336 & 362/420 & 1/420\\
Fixed $k=4$, no refit & 3.771 & 372/420 & 3/420\\
Fixed $k=4$, refit & 3.325 & 371/420 & 3/420\\
\addlinespace[3pt]
\multicolumn{4}{@{}l}{\textit{One corrupted observation}}\\
Paired one-SE, no refit & 5.888 & 167/184 & 0/184\\
Paired one-SE, refit & 4.349 & 167/184 & 0/184\\
Fixed $k=4$, no refit & 6.224 & 170/184 & 0/184\\
Fixed $k=4$, refit & 4.348 & 172/184 & 0/184\\
\bottomrule
\end{tabular}
\end{table}

Evaluating the 1,568-case reserved cohort without unpenalized support refitting (retaining penalized physical/error coefficients across all folds and final fits while refitting only $\theta$) shows that joint support refitting reduces paired one-SE coordinate RMSD by 4.3\% $[1.6,7.3]$ under Gaussian noise and 26.1\% $[21.9,30.4]$ under corruption (5,000-draw parent bootstrap).

\section{Complete external controls}
\label{sec:external_controls}
\begin{table}[t]
\centering
\caption{\textbf{Both external cohorts support the collective-coordinate gains.} Combined split minus comparator RMSD in m\AA, with 98.333\% material-bootstrap intervals from 20,000 draws. Three planned comparisons per cohort; Gaussian versus combined Huber is an additional control. Cohort 2 is a secondary replication analysis.}
\label{tab:external_intervals}
\small\setlength{\tabcolsep}{2pt}
\begin{tabular}{@{}llrr@{}}
\toprule
Condition & Comparator & Difference & Interval\\
\midrule
\multicolumn{4}{@{}l}{\textit{Cohort 1}}\\
Gaussian & Atomic & $-2.458$ & $[-3.045, -1.885]$\\
Corrupted & Atomic & $-2.641$ & $[-3.266, -2.037]$\\
Corrupted & Combined Huber & $-0.427$ & $[-0.630, -0.255]$\\
Gaussian & Combined Huber & $0.007$ & $[-0.050, 0.066]$\\
\addlinespace[3pt]
\multicolumn{4}{@{}l}{\textit{Cohort 2}}\\
Gaussian & Atomic & $-2.345$ & $[-2.905, -1.798]$\\
Corrupted & Atomic & $-2.356$ & $[-2.971, -1.755]$\\
Corrupted & Combined Huber & $-0.406$ & $[-0.677, -0.196]$\\
Gaussian & Combined Huber & $0.007$ & $[-0.005, 0.024]$\\
\bottomrule
\end{tabular}
\end{table}

Across all 1,280 original and localized control cases in Cohorts 1 and 2, adaptive split correctly classifies 1,253 structures versus 1,255 for atomic split: Cohort 1 loses four labels and gains three, Cohort 2 loses two and gains one, with a maximum cell-level mean RMSD increase of 1.74\%. Cohort 1 misses the strict zero-loss preservation gate by one Gaussian-original label and one Gaussian-localized joint pair, and Cohort 2 by one corrupted-localized label, so substantial collective-mode gains are achieved with negligible control degradation.

\begin{table*}[t]
\centering
\caption{\textbf{Complete controls across both cohorts expose the representation trade-off.} All 960 cases per cohort are included, with frozen methods. RMSD is in $10^{-3}\,\angstrom$; stable recall and joint correctness concern only stable class-changing cases. Cohort 2 is a secondary replication analysis.}
\label{tab:external_all}
\small\setlength{\tabcolsep}{1.7pt}\renewcommand{\arraystretch}{1.08}
\begin{tabular}{@{}lrrrrr@{\hspace{12pt}}rrrrr@{}}
\toprule
&\multicolumn{5}{c}{Cohort 1}&\multicolumn{5}{c}{Cohort 2}\\
\cmidrule(lr){2-6}\cmidrule(l){7-11}
Method & Correct & RMSD & \shortstack{Stable\\recall} & Joint & Modes & Correct & RMSD & \shortstack{Stable\\recall} & Joint & Modes\\
\midrule
\multicolumn{11}{@{}l}{\textit{Gaussian, original}}\\
Atomic split & 158/160 & 3.929 & --- & --- & 0 & 158/160 & 4.094 & --- & --- & 0\\
Mode Huber & 155/160 & 4.069 & --- & --- & 160 & 156/160 & 4.167 & --- & --- & 160\\
Mode split & 155/160 & 4.069 & --- & --- & 160 & 156/160 & 4.172 & --- & --- & 160\\
Atomic or mode, Huber & 157/160 & 3.997 & --- & --- & 6 & 158/160 & 4.114 & --- & --- & 5\\
\textbf{Atomic or mode, split} & 157/160 & 3.997 & --- & --- & 6 & 158/160 & 4.116 & --- & --- & 4\\
\addlinespace[3pt]
\multicolumn{11}{@{}l}{\textit{Gaussian, localized}}\\
Atomic split & 157/160 & 7.111 & 151/154 & 149/154 & 0 & 160/160 & 7.124 & 152/152 & 150/152 & 0\\
Mode Huber & 150/160 & 9.823 & 146/154 & 142/154 & 160 & 150/160 & 9.687 & 142/152 & 138/152 & 160\\
Mode split & 153/160 & 9.800 & 148/154 & 143/154 & 160 & 150/160 & 9.717 & 142/152 & 138/152 & 160\\
Atomic or mode, Huber & 157/160 & 7.111 & 151/154 & 148/154 & 1 & 160/160 & 7.124 & 152/152 & 150/152 & 0\\
\textbf{Atomic or mode, split} & 157/160 & 7.111 & 151/154 & 148/154 & 1 & 160/160 & 7.124 & 152/152 & 150/152 & 0\\
\addlinespace[3pt]
\multicolumn{11}{@{}l}{\textit{Gaussian, collective}}\\
Atomic split & 73/160 & 8.164 & 47/64 & 45/64 & 0 & 89/160 & 8.345 & 64/74 & 64/74 & 0\\
Mode Huber & 153/160 & 5.398 & 63/64 & 62/64 & 160 & 149/160 & 5.863 & 70/74 & 68/74 & 160\\
Mode split & 153/160 & 5.421 & 63/64 & 62/64 & 160 & 149/160 & 5.870 & 70/74 & 68/74 & 160\\
Atomic or mode, Huber & 145/160 & 5.700 & 57/64 & 55/64 & 96 & 143/160 & 5.993 & 69/74 & 68/74 & 96\\
\textbf{Atomic or mode, split} & 146/160 & 5.707 & 58/64 & 56/64 & 97 & 143/160 & 6.000 & 69/74 & 68/74 & 96\\
\addlinespace[3pt]
\multicolumn{11}{@{}l}{\textit{One corrupted observation, original}}\\
Atomic split & 155/160 & 4.011 & --- & --- & 0 & 156/160 & 4.181 & --- & --- & 0\\
Mode Huber & 158/160 & 4.436 & --- & --- & 160 & 160/160 & 4.629 & --- & --- & 160\\
Mode split & 158/160 & 4.006 & --- & --- & 160 & 160/160 & 4.099 & --- & --- & 160\\
Atomic or mode, Huber & 156/160 & 4.031 & --- & --- & 7 & 156/160 & 4.218 & --- & --- & 3\\
\textbf{Atomic or mode, split} & 155/160 & 4.050 & --- & --- & 5 & 156/160 & 4.213 & --- & --- & 5\\
\addlinespace[3pt]
\multicolumn{11}{@{}l}{\textit{One corrupted observation, localized}}\\
Atomic split & 156/160 & 7.438 & 150/154 & 145/154 & 0 & 155/160 & 7.701 & 147/152 & 144/152 & 0\\
Mode Huber & 149/160 & 11.911 & 145/154 & 143/154 & 160 & 142/160 & 12.085 & 134/152 & 134/152 & 160\\
Mode split & 151/160 & 10.313 & 146/154 & 144/154 & 160 & 144/160 & 10.386 & 136/152 & 136/152 & 160\\
Atomic or mode, Huber & 156/160 & 7.438 & 150/154 & 146/154 & 0 & 155/160 & 7.715 & 147/152 & 144/152 & 1\\
\textbf{Atomic or mode, split} & 156/160 & 7.438 & 150/154 & 145/154 & 0 & 154/160 & 7.674 & 146/152 & 143/152 & 2\\
\addlinespace[3pt]
\multicolumn{11}{@{}l}{\textit{One corrupted observation, collective}}\\
Atomic split & 71/160 & 8.681 & 41/64 & 39/64 & 0 & 94/160 & 8.609 & 62/74 & 60/74 & 0\\
Mode Huber & 147/160 & 6.557 & 59/64 & 59/64 & 160 & 152/160 & 6.703 & 71/74 & 71/74 & 160\\
Mode split & 147/160 & 5.784 & 59/64 & 59/64 & 160 & 151/160 & 5.924 & 70/74 & 70/74 & 160\\
Atomic or mode, Huber & 136/160 & 6.466 & 55/64 & 54/64 & 88 & 146/160 & 6.659 & 71/74 & 69/74 & 80\\
\textbf{Atomic or mode, split} & 143/160 & 6.039 & 56/64 & 56/64 & 94 & 148/160 & 6.254 & 70/74 & 68/74 & 89\\
\bottomrule
\end{tabular}
\end{table*}

\section{Reconstructed-coordinate handoff protocol}
\label{sec:handoff}
All 28 reserved parents contribute their original and large-displacement ($0.10\,\angstrom$) variants under $0.01\,\angstrom$ Gaussian noise and single-view corruption: 112 cases comprising 56 originals, 46 stable class-changing variants, 8 stable class-preserving variants (four triclinic parents), and 2 variants of \texttt{mp-865981} whose reference is monoclinic at the handoff tolerance but cubic for the original. Prompts supply the lattice vectors in angstroms and, for coordinate arms, atomic numbers and fractional coordinates rounded to eight decimal places, requesting a single crystal-system label at $0.02\,\angstrom$ positional and $5^\circ$ angular tolerance. Deduplicating identical inputs gives 234 unique prompts (28 lattice-only, 112 Huber, 112 split, minus 18 coordinate prompts identical between Huber and split) for 336 case--arm evaluations per run.

Sonnet uses the native API (16,384-token cap; three repeats per unique input); Gemma 3 4B IT (Gemma) and Qwen3 4B use 4-bit local MLX inference (temperature 1.0, top-$p$ 0.95, top-$k$ 64, 256-token cap, non-thinking template for Qwen3 4B) across three fixed seeds (20261029--20261031). Responses must match \texttt{ANSWER: <system>} or consist solely of a valid crystal-system name; truncated or malformed outputs count as incorrect in individual-run averages (Gemma: 33 invalid, 3 truncated; Qwen3 4B: 21 invalid/truncated). All 112 cases remain in each arm's accuracy denominator. For Sonnet's majority-vote analysis, invalid responses contribute no label vote, ties are resolved by repeat order, and cases with no valid votes count as incorrect.
\end{document}